\documentclass{clv3}

\usepackage{hyperref}
\usepackage{xcolor}
\usepackage{natbib}
\usepackage{microtype}
\usepackage{xspace}
\usepackage[utf8]{inputenc}

\usepackage{graphicx}
\usepackage{caption}
\usepackage{subcaption}
\usepackage{booktabs}
\usepackage{tabularx}
\usepackage{todonotes}

\newcommand{\showgbt}{1}

\newcommand{\gbt}[1]{\ifthenelse{\equal{\showgbt}{1}}{\todo{{\bf Gemma says:}\ \textit{#1}}}}
\newcommand{\seb}[1]{\ifthenelse{\equal{\showgbt}{1}}{\todo{{\bf Sebastian says:}\ \textit{#1}}}}
\newcommand{\abhi}[1]{\ifthenelse{\equal{\showgbt}{1}}{\todo{{\bf
        Abhijeet says:}\ \textit{#1}}}}

\newcommand{\niid}{\textit{NotInst-inClass}\xspace}
\newcommand{\nir}{\textit{NotInst-global}\xspace}
\newcommand{\unid}{\textit{Union-inClass}\xspace}
\newcommand{\unr}{\textit{Union-global}\xspace}
\newcommand{\itoi}{\textit{Inst2Inst}\xspace}
\newcommand{\inv}{\textit{Inverse}\xspace}
\newcommand{\union}{\textit{Union}\xspace}
\newcommand{\notinst}{\textit{NotInst}\xspace}

\newcommand{\inclass}{\textit{inClass}\xspace}

\newcommand{\per}{\textit{person}\xspace}
\newcommand{\loc}{\textit{location}\xspace}
\newcommand{\obj}{\textit{object}\xspace}
\newcommand{\art}{\textit{artifact}\xspace}
\newcommand{\com}{\textit{communication}\xspace}
\newcommand{\act}{\textit{act}\xspace}
\newcommand{\oth}{\textit{other}\xspace}

\newcommand{\nnone}{NN-1HL\xspace}
\newcommand{\nntwo}{NN-2HL\xspace}

\newcommand{\conc}{\textit{Conc}\xspace}
\newcommand{\diff}{\textit{Diff}\xspace}
\newcommand{\eg}[1]{\textit{#1}\xspace}

\newcommand{\csshort}{\textsc{CS}}

\issue{1}{1}{2017}
\runningtitle{Instantiation}
\runningauthor{A. Gupta, G. Boleda, S. Pad\'o}

\begin{document}

\title{Instantiation}


\author{Abhijeet Gupta\thanks{E-mail: abhijeet.gupta@ims.uni-stuttgart.de}}
\affil{IMS, Stuttgart University}

\author{Gemma Boleda\thanks{E-mail: gemma.boleda@upf.edu}}
\affil{Universitat Pompeu Fabra}

\author{Sebastian Padó\thanks{E-mail: sebastian.pado@ims.uni-stuttgart.de}}
\affil{IMS, Stuttgart University}

\maketitle


\begin{abstract}
  In computational linguistics, a large body of work exists on
  distributed modeling of \emph{lexical relations}, focussing largely
  on lexical relations such as hypernymy (\emph{scientist -- person})
  that hold between two categories, as expressed by common nouns. In
  contrast, computational linguistics has paid little attention to
  entities denoted by proper nouns (\emph{Marie Curie, Mumbai,
    \dots}). These have investigated in detail by the Knowledge
  Representation and Semantic Web communities, but generally not with
  regard to their linguistic properties.

  Our paper closes this gap by investigating and modeling the lexical
  relation of \emph{instantiation}, which holds between an
  entity-denoting and a category-denoting expression (\emph{Marie
    Curie -- scientist} or \emph{Mumbai -- city}). We present a new,
  principled dataset for the task of instantiation detection as well
  as experiments and analyses on this dataset. We obtain the following
  results: (a), entities belonging to one category form a region in
  distributional space, but the embedding for the category word is
  typically located outside this subspace; (b) it is easy to learn to
  distinguish entities from categories from distributional evidence,
  but due to (a), instantiation proper is much harder to learn when
  using common nouns as representations of categories; (c) this
  problem can be alleviated by using category representations based on
  entity rather than category word embeddings.
\end{abstract}

\section{Introduction}
\label{sec:intro}

A fundamental ontological distinction in language and cognition is
that between \emph{categories} and \emph{entities}: Categories are
equivalence classes that help us handle entities in the world as we
perceive and conceptualize them (objects, people, events, \dots), for
instance by predicting the behavior of new entities based on the
category we assign it to
\cite{murphy02:_big_book_concep,neelakantan2015inferring}.  For
example, the entity \emph{Emmy Noether} instantiates the
\eg{scientist} category, and consequently we expect her to be
associated with academic institutions, to have publications, etc. In
language, entities are typically expressed by proper nouns, and
categories by common nouns. Copular sentences such as \emph{Noether is
  a scientist} arguably express \emph{instantiation}, the most
prominent relationship between entities and categories.



This paper presents a distributional exploration of instantiation.
Instantiation, and entities in general, have received little attention
in linguistics and computational linguistics compared to categories
and relations among categories such as hypernymy (see below and
Section~\ref{sec:relwork} for background).
In this article, we ask the question \textit{Can generic distributed
  representations be used to model instantiation?} and address it with
analysis and experiments. Our ultimate research goal is to advance the
understanding of the notions of entity vs.\ category and their
relation to language; the immediate goals of this article are to test
the hypotheses that 1) entities and categories are represented
differently in distributional space, and 2) not only the broad
distinction between entities and categories, but also instantiation
proper, is recoverable from distributional representations.

\paragraph{Background on Instantiation} In formal semantics ---
concretely model-theoretical semantics and type theory --- common
nouns are modeled as predicates, that is, sets of entities. The
entities are referents in the universe or the utterance
situation~\cite{gamut91:_logic_languag_meanin}.\footnote{This is
  actually the case for nouns with one argument. Relational arguments
  like \eg{father} are typically modeled as sets of entity pairs.}
Proper nouns refer to individual entities.\footnote{There is a
  philosophical discussion about the status of definite descriptions
  and generalized quantifiers; see
  \cite{textor11:_frege_sense_refer}.} From this perspective,
instantiation is simply the set inclusion relationship, and
traditionally instantiation has been assumed rather than investigated:
For example, it is postulated that, in a given model of the world,
Jane Goodall belongs to the set denoted by \eg{scientist} (that is,
Jane Goodall instantiates the category \eg{scientist}), but no attempt
is made at establishing the conditions under which a certain entity
instantiates a category.\footnote{For certain predicates, such as
  vague adjectives, this issue has been thoroughly
  investigated~\cite[and subsequent work]{kamp75}, but not so much for
  category-forming predicates such as nouns -- though
  see~\citet{sassoon2013vagueness}.}  In turn, lexical semantics has
mostly been concerned with category-denoting words and relations
between them, such as synonymy, hypernymy or
meronymy~\cite{cruse1986lexical,Geeraerts2010}, and has traditionally
not focused much on entities. Hypernymy and its mirror hyponymy are
the relationship between two common nouns, one denoting a subcategory
of the other: For instance, \emph{scientist} is a hyponym of
\emph{person}, and \emph{person} a hypernym of \emph{scientist}. In
this case, like in instantiation, one is more specific than the other;
unlike in instantiation, however, both denote categories.

In Knowledge Representation (KR), as a field of Artificial
Intelligence, the situation is different, since KR is interested in
modelling any kind of knowledge. A prominent approach in Knowledge
Representation is Description Logics (DL), a fragment of first-order
predicate logic specifically designed to model the structure of
typical world knowledge \citep{baader2003description}. Central
foundations in DL are classes (analogous to categories), roles, and
individuals (analogous to entities). DL actually distinguishes two
types of knowledge: terminological knowledge, as captured in a typical
ontology, i.e., knowledge about the relationships among classes. This
includes \textit{is-a}, which is like hyponymy. On the other side
there is assertional knowledge, that is, knowledge about relationships
between individuals, on the one hand, and roles and classes, on the
other. Instantiation belongs to the latter, and corresponds to the
\textit{type} relationship in RDF, a representation framework commonly
used in the Semantic Web~\citep{hitzler2009foundations}. Traditional
knowledge bases (Cyc: \citet{Lenat:1995}; SUMO: \citet{Niles2001})
concentrate on terminological knowledge. Current knowledge bases
(FreeBase, \citet{bollacker2008freebase}; YAGO:
\citet{suchanek2007yago}) tend do be rather short on terminological
knowledge and see themselves mostly as repositories of assertional
knowledge about large numbers of entities (although instantiation is
sometimes included under the name of 'entity type',
\citet{neelakantan2015inferring}). WordNet~\citep{Fellbaum-WordNet-1998}
has a kind of hybrid status: It started as a terminological resource
(with the difference to Knowledge Representation resources that it was
structured around word senses), but was later on extended with
knowledge about entities with a new relation,
\textit{instance\_hypernym}, that encodes
instantiation~\cite{Alfonseca2002}.

Most work on distributional semantics has followed traditional lexical
semantics in its focus on categories (mostly common nouns, but content
words in general), examining semantic similarity and relatedness as
well as lexical semantic relations such as hypernymy, besides other
conceptual aspects of meaning like selectional
restrictions~\cite{Baroni2010,baroni2011we,erk2010flexible,roller-erk-boleda:2014:Coling,levy-EtAl:2015:NAACL-HLT}. In
recent years, a few studies have started examining
entities~\cite{herbelot2015mr,herbelot2015building,gupta2015distributional},
and an initial study of our own~\cite{boleda2017instances}
investigated the modeling of instantiation with distributional
methods.

To sum up, previous work either has traditionally not included
instantiation in its purview (formal, lexical, and distributional
semantics) or has tackled it from a rather applied perspective. To address this situation, we make the following contributions:
\begin{itemize}
\item A dataset to evaluate the ability of computational models to
  capture instantiation, analogous to previous datasets built around
  lexical relations such as synonymy and
  hypernymy~\citep{baroni2011we,roller-erk-boleda:2014:Coling}.
\item Experiments on this dataset suggesting that when using generic
  word embeddings, 1)~entities and categories are represented in
  different regions of distributional space, 2)~instantiation is hard
  but largely recoverable from distributed representations using
  nonlinear models, 3)~it is easier to model instantiation using
  entities as a basis for category representation than using common
  nouns, which may have implications for the understanding of the
  relationship between language and cognition.
\item On the methodological level, we show that memorization
  issues that had been previously shown to affect learning in lexical
  semantic relations such as
  hypernymy~\citep{roller-erk-boleda:2014:Coling,levy-EtAl:2015:NAACL-HLT}
  are also an issue with instantiation, and build the dataset so as to
  avoid these issues.
\end{itemize}
Compared to our previous pilot study on this topic
\citep{boleda2017instances}, the current article is extended as
follows: (a), the dataset\footnote{The dataset is available at
  \url{http://www.ims.uni-stuttgart.de/data/Instantiation.html}.}
is substantially improved: we add information about ontological
classes for analysis and confounder selection purposes
(Section~\ref{sec:positive-datapoints}), we add a more challenging set
of confounders (Section~\ref{sec:confounders}), and our data splits
prevent models form performing well simply by memorizing
(Section~\ref{sec:memorization})); (b), we present a detailed
exploration of the properties of categories and entities in the
embedding space (Section~\ref{sec:analysis}); (c), our experiments
include a wider range of models (Section~\ref{sec:exp1-concept}) and
investigate an alternative method for representing categories
(Section~\ref{sec:exp2-centroid})\footnote{The models will be made
  available at publication.}.

\section{A Dataset for Instantiation}
\label{sec:dataset}

Since there is no previous dataset on instantiation (with the
exception of our own pilot study~\cite{boleda2017instances}), our
first contribution is the creation of a benchmark dataset for this
relation with positive and negative examples (word pairs of entities
and categories). For entities, we work with representations of Named
Entities in the public domain such as \eg{Italy} or \eg{George
  Washington} because it is possible to obtain distributed
representations for those using standard methods in distributional
semantics and deep learning. This is common to related work on
entities in distributional
semantics~\cite{herbelot2015mr,gupta2015distributional}.

We follow the procedures of previous studies that created benchmark
datasets for lexical relations
\citep{baroni2011we,roller-erk-boleda:2014:Coling,levy-EtAl:2015:NAACL-HLT},
and extract the positive examples, such as \eg{Virginia Woolf --
  writer}, from WordNet. We pair them with confounders
(\eg{Virginia Woolf -- athlete}) in a principled way, partially
inspired by~\citep{baroni2011we}, to enable a structured investigation
of the different aspects involved in instantiation, in particular the
distinction between instance and category in general vs.\
instantiation proper and the role of similarity.
The next two subsections discuss the positive datapoints and the
confounders, respectively.

\subsection{Positive Datapoints} \label{sec:positive-datapoints}
\begin{figure}[tb]
  \centering
  \caption{  \label{fig:positive-mapping}
Mapping between WordNet synsets and Google News targets.}
  \includegraphics[width=0.8\textwidth]{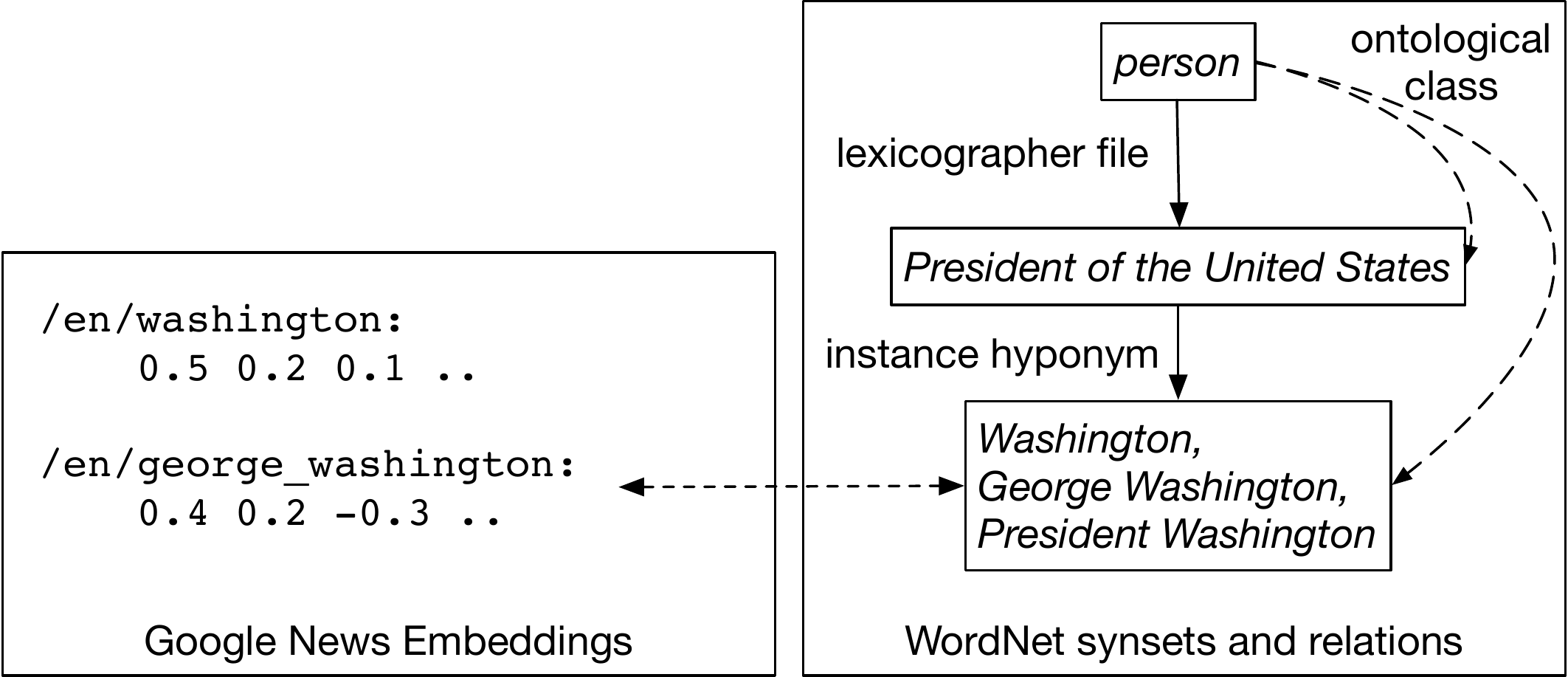}
\end{figure}
Our method for obtaining positive datapoints is shown in
Figure~\ref{fig:positive-mapping}. We start out with WordNet and
extract all pairs of WordNet synsets $(e,c)$ that are linked with the
\textit{instance\_hypernym} relation. This ensures that $e$ is an
entity and $c$ a category. Next, we retain only those pairs for which
we have coverage in a very large distributional resource, the Google
News vector space (\citet{mikolov2013distributed}, see below for
details). The targets for this vector space space were drawn from one
of the largest repositories of entities at the time, the FreeBase
knowledge base~\citep{bollacker2008freebase}. Mapping between WordNet
synsets and FreeBase identifiers requires some heuristics. Notably,
WordNet synsets with multiple elements can map onto several FreeBase
identifiers. For example, the synset \eg{(Washington, George
  Washington, President Washington)} maps onto two FreeBase
identifiers, and thus, two embeddings, that for `Washington' and that
for `George Washington'.\footnote{Some multi-word named
  entities were included in Mikolov et al.'s space as single tokens,
  including `George Washington' (but not the expression `President
  Washington'). Prior to building the space, they lowercased all words
  and combined multi-word expressions such that they were considered a
  single token during space construction; see
  \citet{mikolov2013distributed} for details on how they identified
  multi-word expressions.} We choose the FreeBase identifier that
matches the longest element of the synset, as it will be the least
ambiguous: While \eg{Washington} can refer to different entities (a
person, a state, or a city), \eg{George Washington} almost always
refers to the former president of the USA. Of course the embedding for
\eg{George Washington} will be built from fewer occurrences; our
dataset design favors precision over recall. Note that this strategy
reduces, but does not completely remove, the incidence of short
FreeBase identifiers.

The resulting dataset is a collection of 5,469 positive cases of
instantiation, with 4,750 unique entities and 577 unique categories
(see Table~\ref{tab:positive}). There are more positive word pairs
than entities because some entities in WordNet belong to more than one
category: For example, (\eg{George Washington -- president of the
  United States}), (\eg{George Washington -- general}). There is also
some remaining ambiguity, in the form of different entities with the
same name that belong to different categories -- e.g. \eg{William
  Gilbert the poet} vs.\ \eg{the physicist}. Since these result from
natural properties of instantiation, we perform no further filtering.

Note that there are many more entities than categories, which is to be
expected because there are more entities than (lexicalized)
categories, and some categories contain a large number of
entities. The effect may however be reinforced by the way the Google
News space was built: Its vocabulary consists of the nodes in
FreeBase, a database primarily geared towards entities, though it also
had nodes for categories.

For analysis (see Section~\ref{sec:analysis}) and negative datapoint
selection, we additionally use the WordNet `lexicographer file'
labels, which can be used as proxies for semantic categories or
ontological
classes~\citep{rigau-atserias-agirre:1997:ACL,curran:2005:ACL}.\footnote{If
  the entity and category are assigned different lexicographer files,
  we use the one of the entity; if one of them is missing a class, we
  use the other one. This affects a total of 329 (6\%)
  datapoints. Also, we collapse all ontological classes with fewer
  than 50 occurrences in our data into a class \textit{other}.} These
are shown as rows in Table~\ref{tab:positive} (also see
Figure~\ref{fig:positive-mapping}). Most of the datapoints belong to
ontological classes \per and \loc. The \per class consists of popular
and well known, fictional and non-fictional, historical as well as
modern day people; \loc contains geopolitical entities, like countries
or cities; \obj mostly consists of geographical and natural entities;
\com includes literary texts but also computer programs and operating
systems; \art covers all kinds of man-made entities; finally, \act
consists of famous events.
\begin{center}
\begin{table}
  \caption{\label{tab:positive}Positive datapoints: Statistics
    and examples by ontological class.}
  \begin{tabular}{ l  c  c  c  c  c }
    \toprule
	Ontological Class & \#Entities & Example & \#Categories & Example & \#Datapoints \\ \midrule
\per & 2330 & Madame Curie & 294 & chemist & 2742 \\ 
\loc & 1512 & Oaxaca & 99 & city & 1746 \\ 
\obj & 630 & Nile & 59 & river & 633 \\
\art & 118 & Bastille & 55 & fortress & 121 \\
\com & 97 & Iliad & 43 & epos & 98 \\
\act & 67 & Alamo & 20 & siege & 69 \\
\oth & 60 & Paleocene & 26 & epoch & 60 \\ \midrule
Total unique & 4750 & -- & 577 & -- & 5469 \\ \midrule
	\end{tabular}		
\end{table}
\end{center}

\subsection{Confounders}
\label{sec:confounders}
As mentioned above, we include different types of confounders in our
dataset. In our experiments in instantiation detection
(Sections~\ref{sec:exp1-concept} and~\ref{sec:exp2-centroid}), we ask
the models to distinguish between positive examples and
confounders. More specifically, we generate four sets of confounders
by transforming each entity-category positive example $(e, c)$ as
follows:
\begin{description}
\item[\inv] Swap the positions of entity and category, yielding $(c, e)$.
\item[\itoi] Replace the correct category by a different random
  entity $e'$ of the same ontological class, yielding $(e, e')$.
\item[\nir] Replace the correct category $c$ by a random wrong
category $c''$, from the global distribution of
categories, yielding $(e, c'')$.
\item[\niid] Replace the correct category $c$ by a wrong category $c'$, this time sampling 
from the same ontological class, yielding $(e, c')$.
\end{description}

\inv tests that the models correctly capture the asymmetric nature of
instantiation. \itoi checks that the models are not fooled by
similarity (entities in the same ontological class are similar to each
other, see Section~\ref{sec:analysis}).  Finally, \nir and \niid aim
at testing that models actually learn the relation between a specific
entity and a specific category, as opposed to learning to classify
entities vs.\ categories in general~\citep{levy-EtAl:2015:NAACL-HLT}.
The difference between \niid and \nir is one of difficulty: In \niid,
confounder categories come from the same ontological class as the
correct categories, and thus are semantically more similar to the
correct category (e.g.\ pairing \textit{George Washington} with
\textit{astronomer}) than in \nir (where \textit{George Washington} is
paired with \textit{river}).  Table~\ref{table:confounders} shows two
examples of a positive datapoint with its corresponding confounders.

When pairing confounders with the positive examples, we obtain in four
different balanced subsets, consisting of pairs of expressions for
which the instantiation relationship either holds (positive examples)
or does not (confounders): Pos+\inv, Pos+\itoi, Pos+\nir and
Pos+\niid. Two final variants, Pos+\unid and Pos+\unr, combine the
positive examples with \inv, \itoi, and one of the two
\textit{NotInst} variants, respectively. These two variants are more
challenging in that they require models to distinguish positive
examples from confounders of different types, and have a 1:3
positive-to-negative ratio.
\begin{table}[]
  \label{tab:examples}
\caption{\label{table:confounders}Examples of confounders. POTUS = President of the United States.}
\begin{tabular}{ lll }
  \toprule
   		Type & Example 1 & Example 2  \\ \midrule
   		Positive & George Washington -- POTUS & Mumbai -- city \\
   		\inv & POTUS -- George Washington & city -- Mumbai \\
   		\itoi & George Washington -- Peter Behrens & Mumbai -- Vicksburg \\
   		\nir & George Washington -- river & Mumbai -- statesman \\ 
   		\niid & George Washington -- astronomer & Mumbai -- residential area \\ \bottomrule 		   		
\end{tabular}
\end{table}

\subsection{Dataset Partitioning and Memorization}
\label{sec:memorization}

We split each dataset variant into training, validation and test sets
(80, 10, and 10\% respectively). This is however not enough to make
sure that the models can generalize: the related task of hypernymy
detection has been shown to suffer from the problem of
\textit{memorization}~\cite{roller-erk-boleda:2014:Coling,levy-EtAl:2015:NAACL-HLT},
that is, models learning by heart that certain words (such as
\textit{animal}) make good hypernyms instead of truly learning the
hypernymy relation. The problem is that, in a na\"ive random split of
the datapoints, even if the pairs are not reused across partitions,
individual members of the pair can be. Thus, good results can hide a lack
of generalization, in particular for frequent categories.

To address this issue, we adapt the methodology of
\citet{roller-erk-boleda:2014:Coling} which ensures that there is zero
lexical overlap between training, validation, and test
sets. Specifically, we split the test set into many
equal-sized test folds and remove overlap with the training and
validation data: For example, if \eg{(George Washington, President of
  the United States)} occurs in a test fold, then all pairs containing
either \eg{George Washington} or \eg{President of the United States}
are removed from the corresponding training and validation
data.\footnote{We choose the number of test folds so that the average
  size of the training set after removing the lexical overlap is 90\%
  of the original training data (fewer, and therefore larger, test
  folds lead to more excluded training data). This results in 83 test
  folds.} Leave-one-out evaluation, as chosen by
\citet{roller-erk:2016:EMNLP2016}, would have increased computational
load substantially.

\subsection{Distributional Space} \label{sec:distspace} For the
analysis and experiments below, we represent both entities and
categories in terms of their Google News
embeddings~\citep{mikolov2013distributed}. Mikolov et al.'s space
constitutes, to our knowledge, the largest existing source for entity
embeddings. The space consists of about 1.4 million 1000-dimensional
embeddings whose targets are drawn from FreeBase
\citep{bollacker2008freebase}. The embeddings were extracted from a
100 billion word Google news dataset using the Skip-gram
algorithm~\citep{mikolov2013distributed}. As mentioned above, the vast
majority of these entries are entities; however, the space also models
category-denoting nouns. By the definition of our data selection
strategy, the space contains, for each entity, the category it
instantiates. We rescale the embedding values column-wise so they
lie within the [-1, 1] range.

We treat the embeddings as static, rather than optimizing them for the
task at hand, because we are interested in the geometric structures
present in a generic word meaning space, following the rationale of
the Distributional Memory~\citep{Baroni2010}. Moreover, optimizing the
vectors might overfit on the instantiation relation on our relatively
small dataset.

\section{Exploratory Data Analysis}
\label{sec:analysis}

We perform exploratory data analysis on the positive portion of our
dataset (Section~\ref{sec:positive-datapoints}) to understand the
properties of entities vs.\ categories in distributional space.  We
ask the following questions: 1) Are entities and categories
represented distinctly in distributional space? (following up on
\citet{boleda2017instances}); and, 2) What is the relationship between
entities and their corresponding categories?

\subsection{Instances and Concepts in Space}

\begin{figure}[!t]
  \centering
  \caption{Exploratory data analysis}          
  \begin{subfigure}{0.97\linewidth}
 \caption{\label{fig:pca_byTypes}Entities and categories in
          distributional space, first two PCA dimensions.          
        }
\includegraphics[width=0.9\textwidth]{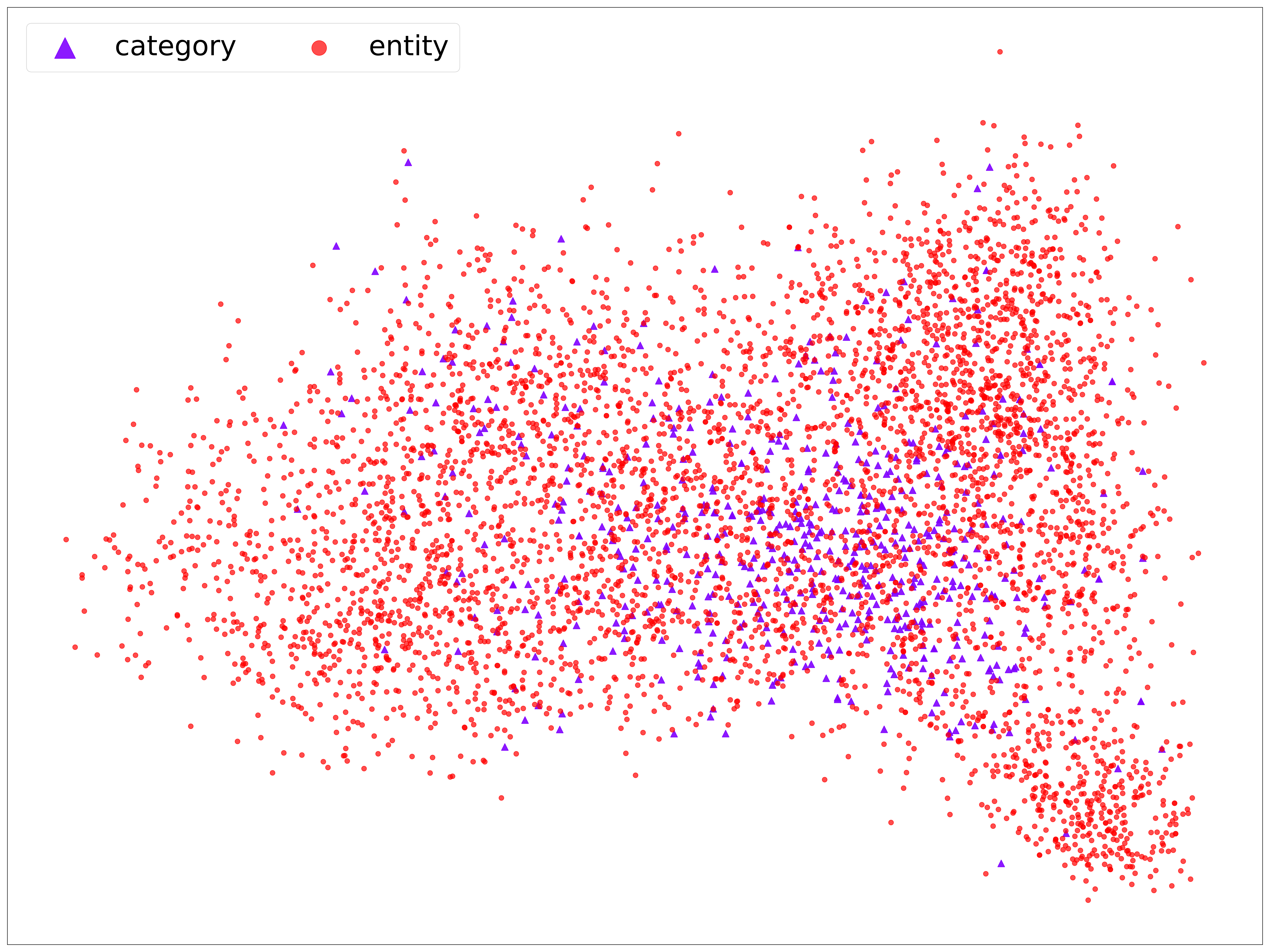}    
 \end{subfigure}
 \begin{subfigure}{0.97\linewidth}
	\caption{\label{fig:pca_byDomains}Ontological classes}	\includegraphics[width=0.9\textwidth]{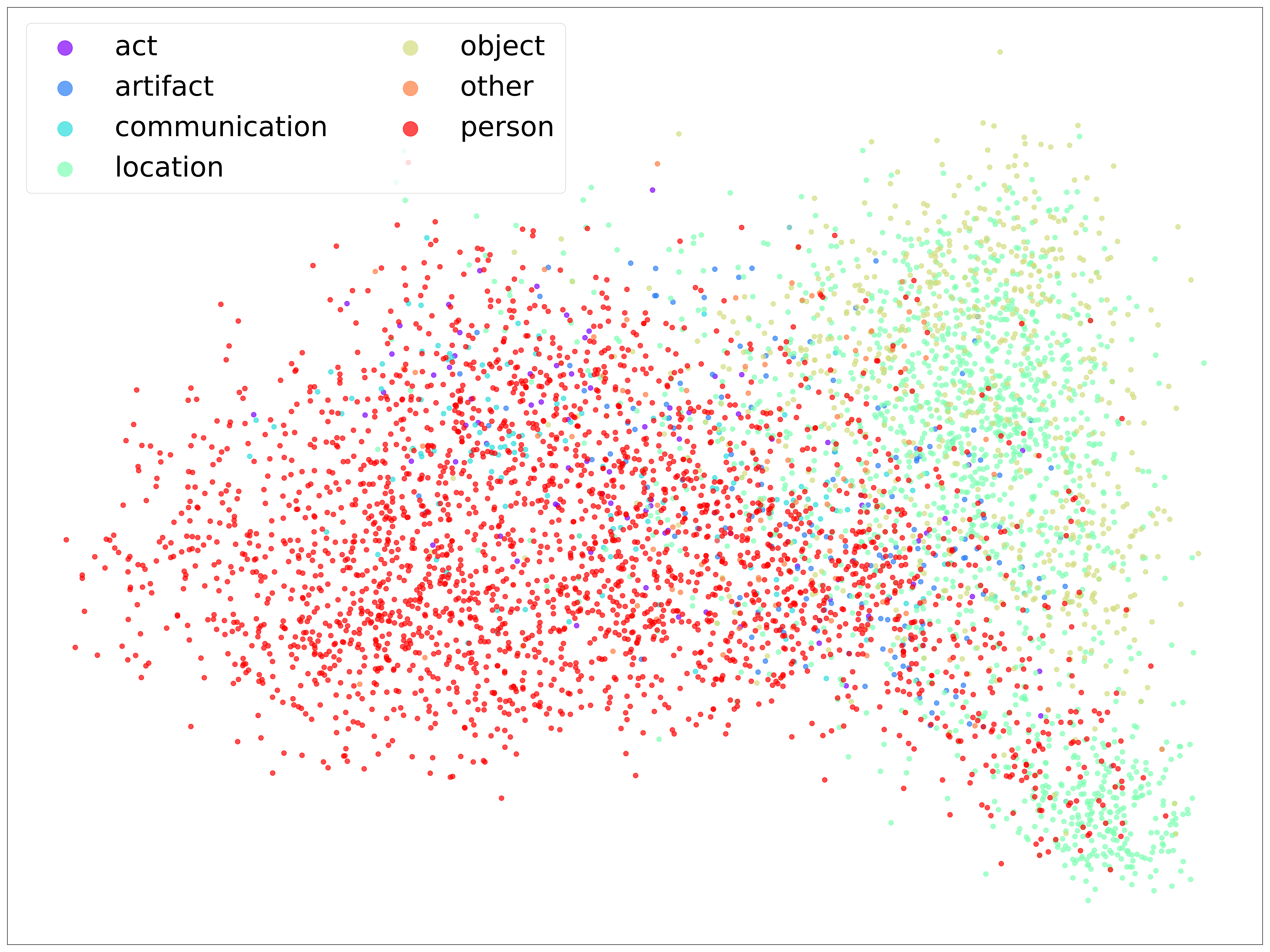}   
 \end{subfigure}
\end{figure}

Figures~\ref{fig:pca_byTypes} and~\ref{fig:pca_byDomains} represent
the first two dimensions of a Principal Component Analysis (PCA)
transformation of the original 1000-dimensional embeddings.
Figure~\ref{fig:pca_byTypes} shows that, while entities and categories
are not in distinct (or even linearly separable) regions of this
reduced space, categories (purple triangles) are concentrated in the
middle region of the graph, creating a radial structure. This constitutes
initial evidence of the distinction between entities and categories in
distributional space.

Figure~\ref{fig:pca_byDomains} shows that the data is also sensitive
to ontological distinctions such as those encoded in WordNet (see
Table~\ref{tab:positive}, Section~\ref{sec:dataset}).
The figure shows a clear division between
animate (\per, left) and inanimate entities/categories 
(\loc and \obj, right; although these two
classes largely overlap, \obj is more concentrated in the
upper right part and \loc in the lower right part).
In the middle, partially overlapping with the classes above, we find the smaller 
classes \art, \act (human-centered events), and 
\com, as well as the catch-all \oth class.

\begin{figure}[!h]
  \centering
  \caption{Representative clustering solutions.}
  \begin{subfigure}{0.86\linewidth}
  \caption{\label{fig:histogram_6solution_ClusterTypes} Distribution of entities vs.\ categories.}
\includegraphics[width=\textwidth]{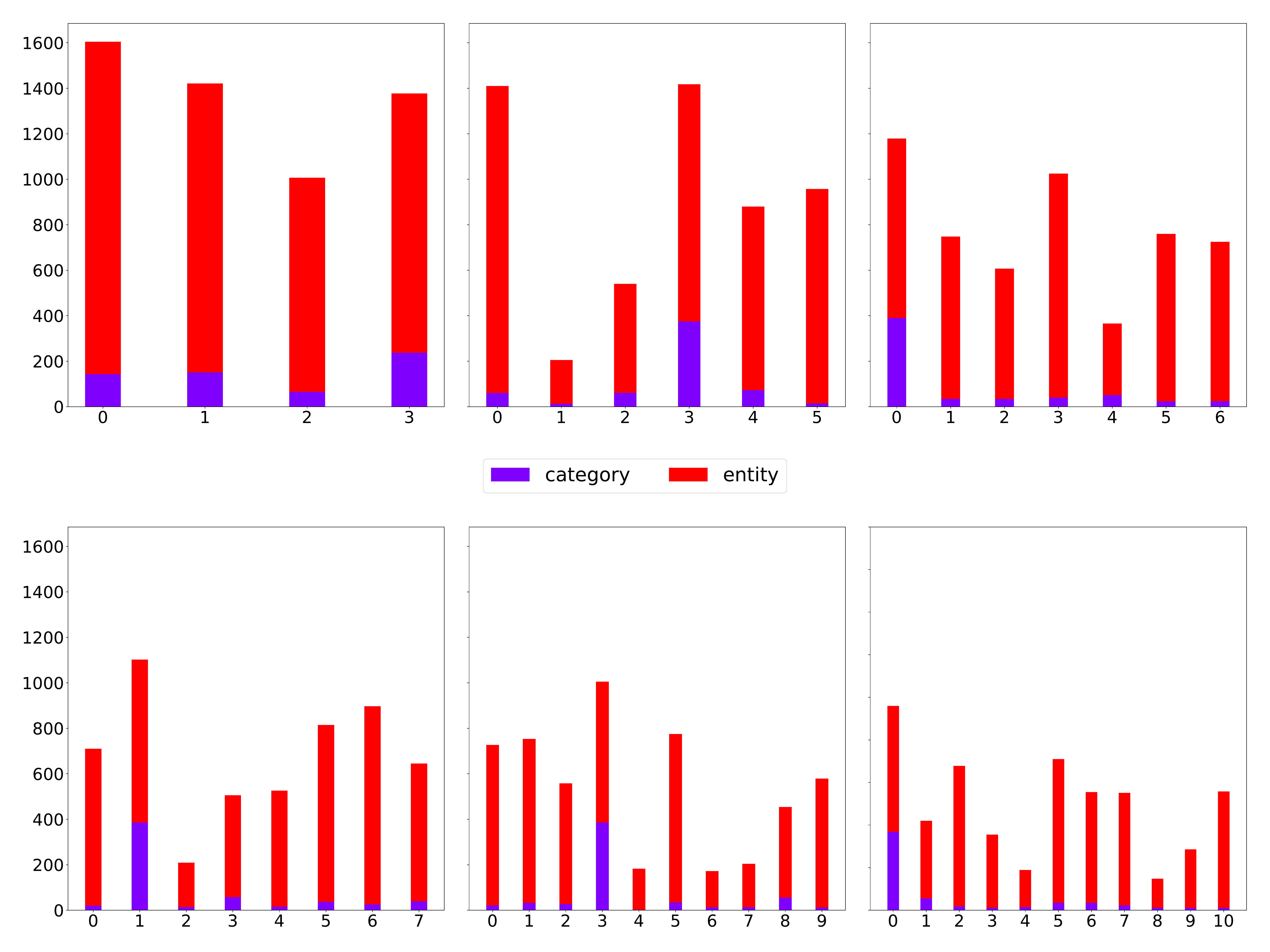}    
  \end{subfigure}
\begin{subfigure}{0.86\linewidth}
\caption{\label{fig:histogram_6solution_ClusterDomains} Distribution of WordNet ontological classes.}
\includegraphics[width=\textwidth]{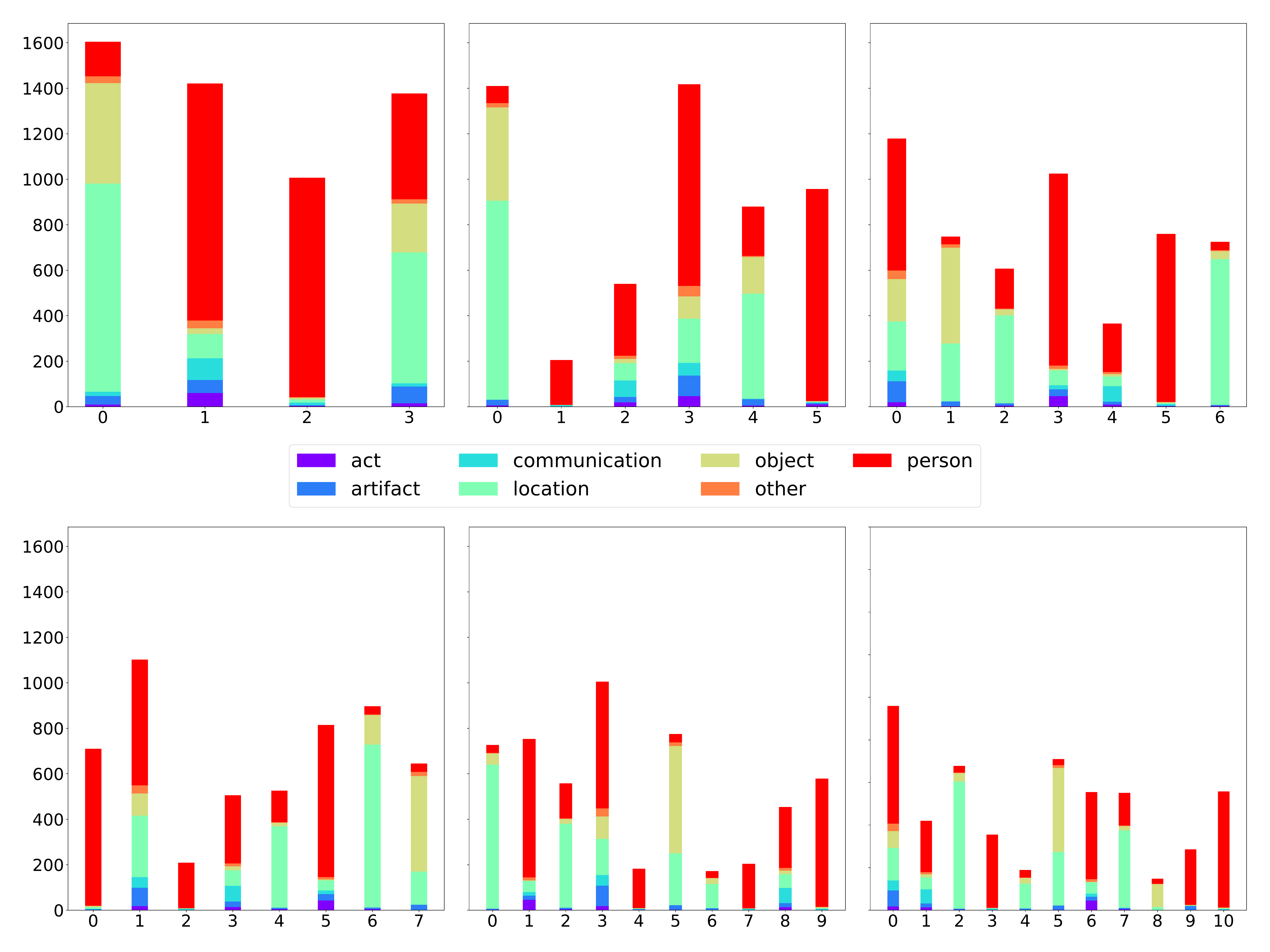}
\end{subfigure}
\end{figure}

A clustering analysis supports these findings. We use the standard
\emph{$k$-means} algorithm to compute clustering solutions with 2 to
15 clusters and analyze results against the entity/category axis as
well as WordNet ontological classes.\footnote{Since the centroid
  seeds are randomly initialized, the algorithm is run 10 times with
  different seeds and the model with the best performance in terms of
  inertia is selected.  The number of iterations for each model run is
  set to a maximum of 10000 with convergence at tolerance value of
  1.0.} Figure~\ref{fig:histogram_6solution_ClusterTypes} shows a
representative selection of clustering solutions
along with the
distribution of entities and categories for each of the clusters
within a solution. They range from 4 clusters at the top left to 11
clusters at the bottom right.

The figure shows a similar trend to Figure~\ref{fig:pca_byTypes}: On
the one hand, the clustering algorithm does not use the division
between entities and categories as its primary organizing principle
for cluster solutions, especially not with a small numbers of clusters
(we will see below that it uses the animate/inanimate division); on
the other, for solutions with a higher number of clusters, categories
tend to group together in a single cluster (cluster 3 in clustering
solution 6 or \csshort6 as well as \csshort10, cluster 0 in \csshort7
and \csshort11, cluster 1 in \csshort8). This is consistently found in
all solutions with higher number of clusters (not shown). This
`conceptual cluster' is mainly a concentration of \per categories
which reflect professional/societal roles, like \textit{musician,
  physicist, minister, king, environmentalist, engineer, artist},
etc. Note that this cluster also contains many entities; they are
related to these categories.

\begin{table}[t]
\centering
\caption{\label{tab:domain_division}Prominent themes in
  representative clustering solutions (from manual analysis).}
\begin{tabular}{cp{5cm}ll}
\toprule
  Clustering & \per & \loc & \obj \\ 
	\midrule
 6 clusters & 0: Political, Academics, & 2: American & 2: American \\ 
                & Science and other professions  & & \\
	& 3: Historical and Religious & 6: Eurasia & 6: Eurasia \\
	& 4: Writers & & \\
	& 5: Musicians, Composers and Painters & & \\
	\midrule	
	11 clusters & 0: Political and Science & 2: America & 4: Astronomy\\
                & 1: Composers & 7: Eurasia & 5: Larger Geography - \\
	& 3: Writers & & (Seas, Oceans, Islands,\\ 
	& 4: Astronomers & & Continents, Countries) \\
	& 6: Historical and Religious & & 8: Small Geography -\\
	& 7: Explorers & & (Rivers, Canals, Parks)\\
	& 9: Painters & & \\
	& 10: Musicians, other professions & & \\
	\midrule
\bottomrule
\end{tabular}
\end{table}

Figure~\ref{fig:histogram_6solution_ClusterDomains} presents the same
6 clustering solutions, this time showing the distribution of the
WordNet ontological classes. As could be expected, the
animate/inanimate distinction appears prominently: Clusters 0 and 3 in
\csshort4 contain most \obj and \loc datapoints, whereas clusters 1
and 2 contain most \per datapoints. However, while the \per/\loc/\obj
distinction is relevant throughout the clustering solutions, we do not
see a clear mapping between clusters and ontological classes in
general. Manual inspection reveals the relevance of topical
distinctions, with clusters partitioned by differences like American
vs.\ Eurasian locations (see Table~\ref{tab:domain_division}).

To sum up, both the PCA reduction and the clustering analysis suggest
the primacy of ontological and topical dimensions in distributional
space, at the same time suggesting a role for the entity/category
axis, as categories tend to cluster together (also see next subsection).

\subsection{Categories: Concept vs.\ Centroid-based
  Representation} \label{sec:conceptandcategoryreps} The above
analysis suggests that entities and categories live in somewhat
different parts of the space and that there is no simple relationship
between entities and the categories they instantiate. This observation
motivates a departure from a \textit{concept}-based representation of
categories based on the occurrences of the noun denoting the category
(e.g.,\ \textit{scientist}). We experiment with an alternative
representation of categories, namely the \emph{centroid} of the
entities that instantiate the category: \textit{Madame Curie,
  Einstein, Lavoisier, Mendel}, etc. A centroid is a dimension-wise
average vector, so if computed over entities of the same category
it can be expected to represent that category,
smoothing out idiosyncracies of particular instances. This
representation resonates with two classical models in Cognitive
Science: On the one hand, exemplar models \citep{Nosofsky1986},
because it bases categories on specific instances; on the other,
prototype theory~\cite{rosch}, as the centroid can also be seen as a
prototype because of the ``smoothing out'' effect (see
\citet{murphy02:_big_book_concep} for discussion). Both have
demonstrated desirable properties in Machine Learning and NLP studies
\citep{jaekeletal08,reisinger-mooney:2010:NAACLHLT}.

\begin{table}[]
  \caption{\label{tab:similarities} Cosine similarities across and
    within categories (means and standard deviations). ``Across-categories'' compares an entity/category to all other
    entities/categories: \eg{Madame Curie} vs.\ \eg{George Washington}, \eg{Einstein},
    \eg{Mumbai}, \eg{Nile}, etc., and \eg{scientist} vs.\
    \eg{president of the United States}, \eg{city}, \eg{river}, etc.
    ``Within-categories'' restricts comparison to a single category:
    \eg{Madame Curie} vs.\ \eg{Einstein}, \eg{Mendel}, etc.}
  \centering
  \begin{tabular}{lrr}
    \toprule
    & Across-categories & Within-categories\\
    \cmidrule{1-3}
    Entities   &  0.05 (0.07) & 0.22 (0.11) \\
    Categories    &  0.06 (0.06) & -- \\
    Centroids   &  0.20 (0.12) & -- \\
    \cmidrule{1-3}
    Entity-Category &  0.04 (0.05) &  0.16 (0.09) \\
    Entity-Centroid &  0.10 (0.09) & 0.55 (0.11) \\
    Category-Centroid &  0.08 (0.07) & 0.29 (0.14)\\
  \end{tabular}  
\end{table}

\begin{figure}[tb]
  \centering
  \caption{\label{fig:person} Entities, centroid embeddings, and
    common noun embeddings for categories \textit{patriarch} and
    \textit{geneticist}.}
\includegraphics[width=0.75\textwidth]{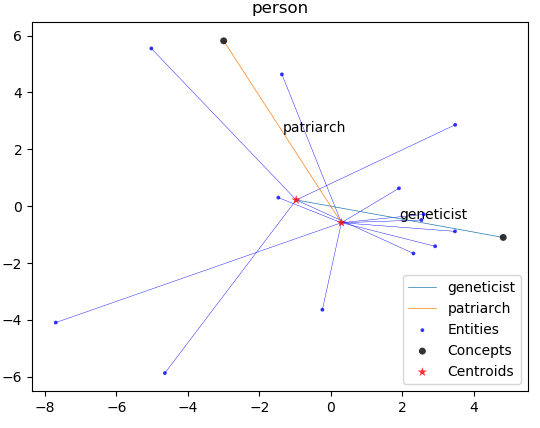}
\end{figure}

For our dataset, Table~\ref{tab:similarities} shows that the
centroid-based representation appears to be promisingly
well-behaved. While the average similarity of each entity to its
category is 0.16 (e.g.\ \eg{Madame Curie} - \eg{scientist}), the
average inter-entity similarity within categories is 0.22 (\eg{Madame
  Curie} - \eg{Einstein}, \eg{Mendel}, \dots).\footnote{The similarity
  to the centroid is 0.55; a high similarity is to be expected here
  simply from the definition of centroid.} This means that
distributional representations of entities belonging to a certain
category are more similar to each other than to the common noun
denoting the category they all instantiate.
Moreover, the fact that the average similarity between centroids and
categories is higher than between entities and their categories (0.29
vs.\ 0.16) suggests that centroids make good category representations, indeed
smoothing out idiosyncratic differences between the entities as
expected. Figure~\ref{fig:person} illustrates this, showing a PCA
reduced representation of categories \textit{patriarch} and
\textit{geneticist}. The
entity embeddings are shown as small dots, linked to their respective
category centroids (shown as red stars). The category embeddings are shown
as large dots. We see that not only are the common noun embeddings
marginal in terms of the overall data distribution, they are in fact
more similar to the other categories' entities than to their own.

These results motivate Experiment~2 in
Section~\ref{sec:exp2-centroid}, where we show that the benefits of
centroid-based representations carry over to the actual task of
instantiation classification.

\section{Experiment 1: Instantiation Detection as Classification} \label{sec:exp1-concept}
We frame instantiation detection as a supervised binary classification
task. The model needs to decide, given an entity-category pair 
as input, whether there is an instantiation relation between the entity 
and the category. 

\subsection{Evaluation} \label{sec:eval-exp1}

We use F$_1$-score on the positive data samples as our evaluation
measure.\footnote{While accuracy would be a simple alternative evaluation
measure on the balanced variants of our dataset (Pos+\niid, Pos+\nir, Pos+\itoi,
Pos+\inv), F$_1$ generalizes well to minority-class
setups like Pos+\unid, Pos+\unr.} We compute F$_1$-scores on individual
folds and micro-average them (the  difference between micro and macro
average is negligible in our case, due to the equal  
number of datapoints in each test fold; see Section~\ref{sec:memorization}).

\subsection{Models} \label{sec:exp1-models}

We test two generic model architectures. The first one is a simple
logistic regression (LR) classifier. The second one is a feed-forward
neural network (NN) classifier.  \footnote{The NN classifiers were
  built with the Keras toolkit, \url{https://keras.io}} The input and
output of both classifiers are identical: for each datapoint (i.e.,\
an \emph{entity--category} pair $(e,c)$), the input is a function of
the two embeddings, $f(\vec e, \vec c)$. For the moment, we only
consider concatenation (\conc):
$f(\vec e, \vec c) = \langle e_1,\dots,e_n,c_1,\dots,c_n\rangle$; in
Section~\ref{sec:results-inputrepresentation} we will consider
the difference between the vectors. The
output is a binary value indicating whether the entity $e$
instantiates the category $c$.

We choose LR and NN for their comparability: The LR classifier has no
hidden layers, thus treating all features as independent, while the NN
model introduces hidden layers that can model non-linear relationships
between input and output and
facilitates interactive behavior between the input features. This
setup allows us to gauge to what extent these aspects affect
instantiation while staying within the same general modeling
framework. For the NN classifier, we experimented with 1-4 hidden
layers but found a decline in model performance from 3 layers
onwards. Hence, we report results only on the architectures with one
hidden layer (\nnone) and two hidden layers (\nntwo).

\paragraph{Model Optimization} We use mean cross-entropy as loss
function and \textit{softmax} as activation function
for the output layer. All
hidden layers use \textit{tanh} as activation function. The number of units in
each hidden layer of the NN models is optimized for each model
separately. We consider the following values: 5, 10, 50--800 (step
size 50).
All models are trained using Adadelta optimization
\citep{zeiler2012adadelta} to a maximum of 2000 epochs with early
stopping (we found that models typically converge at 50--100 epochs).
To reduce overfitting, we introduce a dropout layer in front of each
hidden layer with a standard dropout value of 0.5
\citep{baldi13:_under_dropout}.\footnote{We experimented with
  additional L$_2$ weight regularization but did not find any
  benefits.}

\paragraph{Baselines} We consider two baselines. A
 frequency baseline (BL$_{\text{freq}}$) assigns the
positive class randomly with the true class probability (50\%
for the balanced variants and 25\% for the union variants). We do not
consider a most frequent class baseline, because this baseline would
not make any positive predictions on the union variants, where the
negative class is dominant. Instead, we consider a baseline that
\textit{always} assigns the positive class (BL$_{\text{pos}}$). This
baseline should show strong results in our evaluation scheme (F$_1$
score on positive class) since it always yields a perfect recall.
\subsection{Main results} \label{sec:exp1-main-results} 
\begin{center}
\begin{table}[tb]
	\caption{\label{table:main_results} F$_1$ scores for instantiation detection, concatenated embeddings.}
	\begin{tabular}{ l  c c  c  c  c }	
	Dataset & BL$_{\text{freq}}$ & BL$_{\text{pos}}$ & LR & \nnone & \nntwo \\ \midrule 
	Pos + \inv & 0.50 & 0.67 & \textbf{0.96} & \textbf{0.96} & \textbf{0.96} 	\\
	Pos + \itoi & 0.50 & 0.67 & 0.90 & \textbf{0.91} & \textbf{0.91}	\\ 
    Pos + \nir & 0.50 & 0.67 & 0.55 & \textbf{0.85} & 0.82	\\
	Pos + \niid & 0.50 & 0.67 & 0.55 & \textbf{0.70} & 0.69	\\ 
          \midrule
	Pos + \unr & 0.25 & 0.40 & 0.55 & 0.75 & \textbf{0.76}	\\ 
	Pos + \unid & 0.25 & 0.40 & 0.55 & 0.57 & \textbf{0.63}
	\end{tabular}
\end{table}
\end{center}
Table \ref{table:main_results} reports the main results for the three
models and the baselines.
The top part of the table lists results for the balanced dataset
variants, all of which have a positive class baseline of 0.67. The
bottom part shows results for the \union variants, where the positive
class is now a minority class, with a correspondingly lower positive
class baseline of 0.40. The dataset variants are arranged in their
expected level of increasing difficulty in classification from top to
bottom. Indeed, we see a clear trend of decreasing F-Scores. Since the
first two types of negative examples (\inv and \itoi) can be
classified correctly solely by learning to distinguish categories and
entities, we concur with \citet{boleda2017instances} that the
high-level distinction between categories and entities can be made
quite easily with standard embeddings.

In contrast, \notinst (the setup where the confounders are also
\textit{entity--category} pairs, just not ones that exemplify the
instantiation relation, like \eg{Madame Curie - lawyer}) is a quite difficult setup for which the
scores decrease markedly. Thus, it is the presence or absence of the
specific instantiation relationship, over and above the general 'type
signature', that is  difficult to determine.  In line with this
interpretation, we see that for \notinst and \union, the
\textit{inClass} variants, where the confounders are more semantically
similar to the correct answers, are much harder than the
\textit{global} variants, with models outperforming the baseline by
at most 3 points (0.70 vs.\ 0.67).

LR beats the baseline for \inv and \itoi but not for \notinst or the
\union variants.
The failure of LR -- a linear classifier -- to properly learn instantiation is in line with the observations
by \citet{roller-erk:2016:EMNLP2016} and
\citet{levy-EtAl:2015:NAACL-HLT}, who found that linear classifiers
are generally unable to learn semantic relations from vanilla
embeddings.

In contrast, the NN models beat the baseline for all variants, even
though they also see a decrease in performance for the harder
variants. The benefit of the NN architecture compared to LR correlates
strongly with the difficulty of the task: For the easiest \inv and
\itoi variants, LR and NN perform at par, while the hard \notinst
cases see differences of more than 10 and 30 percent, respectively.

The two NN models perform similarly.
On the easiest variants (\inv and \itoi), both models do equally well.
The model with one hidden layer performs better on the balanced
\notinst variants but is beaten by the two-layer model on the \union
variants. This indicates that the combination of different confounders
calls for a model that can perform substantial transformations of the
feature space. Note also that all models perform worse on the \union
variants than the average of their performances on the individual
variants, indicating that the different kinds of confounders call for
different transformations from the input through the hidden layers.
In the remainder of this section we focus on the neural network model with
two hidden layers because it is the best on the hardest variants.

\subsection{Error Analysis} \label{sec:error-anal-classif-exp}
%
%
%

An error analysis of the predictions made by our best model, \nntwo,
shows that most errors stem from
\textit{semantic relatedness}, which is a well known problem of
distributional semantics \citep{Radovanovic:etal:2010b,baroni2011we}. 

More concretely, we find two distinct linguistic
phenomena leading to errors affecting categories and two phenomena
affecting entities. The first problematic phenomenon for categories is
conceptual similarity. For example, \eg{Edna Ferber}, a
\eg{writer}, is also predicted to be a \eg{composer}, probably due to the
similarities between the two artistic occupations (compare the use of
English \textit{to compose} for the production of both text and of
music). The second problematic phenomenon is association.
For example, the \eg{Cheshire Cat} from the book \textit{Alice in
  Wonderland} is correctly recognized as a \eg{fictional character},
but it is also wrongly predicted to be a \eg{writer}, presumably because it is
often discussed in the context of literature and literary theory.

As for entities, the first source of errors is
referential ambiguity: While many names are
unambiguous when given in their full form,
texts often use abbreviated versions that can refer to multiple
entities (remember our earlier \eg{Washington} example: the person, the state, or the city). For
some names, even the full form is ambiguous, like for
\eg{Albert Smith}, the name of various politicians, cricketers and
footballers. In the absence of large-scale reliable co-reference
resolution and entity linking methods, researchers need to resort to
heuristics to map corpus occurrences onto concrete entities, and wrong
decisions result in biased embeddings.
This type of error is analogous to the pervasive issue of ambiguity in
lexical semantics~\cite{cruse1986lexical}, though from a referential perspective.
Occasionally, this problem also spills over into the category domain:
When frequent entity names include the category name (\eg{gulf -- Gulf of
  Patras}), the embedding for the category term \textit{gulf} may be biased by occurrences that are really parts of entity
names.\footnote{Recall from Section~\ref{sec:dataset} that our
  vocabulary is lowercased. For English, this problem could be
  alleviated by not lowercasing, which however introduces large
  amounts of wrong ambiguity. For languages which capitalize all nouns, such as German,
  or logogrammatic writing systems such as Chinese, the problem
  persists even when lowercasing.}

The second source of errors for entities is \textit{changes in
  the world over time}. For example, \eg{Stagira} was a city in
ancient Greece, and is recorded as such in WordNet. It was however not
predicted as such, presumably because the newswire texts underlying
the embeddings only refer to it as a ruins or more generally as a
historical site. Similarly, \eg{Etruria} used to be an independent
\eg{country} in pre-Roman times, but
is not predicted to be a country, probably due to the predominance of
occurrences related to when it was a region of the Roman empire.
Thus, for entities we find a strong effect of referential aspects of
meaning, which are understudied in Computational Linguistics and Linguistics~\cite{mcnally-boleda17}.

Occasionally, we also encountered errors due to missing
  relational information in WordNet. For example, \eg{Richard
  Brinsley Sheridan} was both a \eg{playwright} and a British
member of parliament, i.e., a \eg{politician}. Only the former
relation appears in WordNet, but our model predicts also the second.
Similarly, \eg{Yalta} is only listed as a resort \eg{city} on
the Black Sea in WordNet, but our classifier adds the information that
it is a \eg{port}. These observations are in line with the known
incompleteness of knowledge bases and the usefulness of distributional
methods to complete them \citep{min2013distant}.

\begin{table}[t]
\begin{center}
  \caption{\label{table:input-rep}Effects of input
    function (concatenation vs. difference) on F$_1$ score for best
    model (\nntwo).}
	\begin{tabular}{ l c  c }
          \toprule
	Dataset & Conc & Diff\\  \midrule
	Pos + \inv & 0.96 & \textbf{0.97}\\ 
	Pos + \itoi & \textbf{0.91} & \textbf{0.91}\\ 
	Pos + \nir & \textbf{0.82} & \textbf{0.82}\\ 
          Pos + \niid & 0.69 & \textbf{0.72}\\
          \midrule
	Pos + \unr & \textbf{0.76} & 0.75\\ 
        Pos + \unid & 0.63 & \textbf{0.67}
	\end{tabular}
\end{center}
\end{table}

\subsection{Effect of input representations}
\label{sec:results-inputrepresentation} 

We next test a different function to combine the representations of
the two input elements. Above, we only considered concatenation, which
enables the model to freely combine the information of the two
embeddings, but does not make the dimension-wise correspondence
between them explicit. We now use the difference
function (\diff), defined as
$f(\vec e, \vec c) = \langle e_1-c_1,\dots, e_n-c_n\rangle$.  This
representation, inspired by \citet{mikolov2013distributed} and
\citet{roller-erk-boleda:2014:Coling}, explicitly links the
information in the input pair by dimension. Thus, the difference input
provides the model with a clearer notion of how the category and
entity embeddings are located \textit{relative} to one another, at the
loss of their \textit{absolute} positions in embedding space. Note
also that \conc produces 2,000-dimensional while \diff produces only
1,000-dimensional input embeddings.

As Table~\ref{table:input-rep} shows, the input representation does
not play a major role. Results across variants are generally very
close, with one exception: \diff yields better results for
the \inclass variants, which are the most difficult.
This could be due to an
advantage of the \diff embeddings for the case of highly
similar confounders: While \conc also contains the necessary
information to make the decision, this information is distributed over
components in the embeddings whose correspondences by dimension the
model must recover.

\subsection{Impact of Memorization} \label{sec:results-memorization}
\begin{center}
\begin{table}[tb!]
  \caption{\label{table:memorization}Effects of memorization on F$_1$ score for best model (\nntwo).}
  \centering
	\begin{tabular}{ l  c  c c }
          \toprule
          Dataset  & Filtering & No filtering & $\Delta$ \\
          \midrule
	Pos + \inv & 0.96  & 0.98 & 0.02 \\ 
	Pos + \itoi & 0.91 & 0.99 & 0.08 \\ 
	Pos + \nir & 0.82  & 0.90 & 0.08 \\ 
          Pos + \niid & 0.69  & 0.83 & 0.14 \\
          \midrule
        Pos + \unr & 0.76  & 0.89 & 0.13 \\ 
        Pos + \unid & 0.63 & 0.81 & 0.18
	\end{tabular}
\end{table}
\end{center}
Section~\ref{sec:memorization} discussed our strategy to counteract
possible memorization effects. This section demonstrates that
memorization issues, identified in previous literature for hypernymy
detection~\cite{roller-erk:2016:EMNLP2016}, affect instantiation as well.
Table~\ref{table:memorization} shows the results of Experiment~1 with
our filtering (same as above; left column) and without
counter-memorization filtering (middle column), as well as the
difference in results between the two set-ups (right
column). Recall from above that the counter-memorization filtering
removes all lexical overlap between the test set and the training and
validation sets. Therefore, models are not asked to learn from \eg{Einstein} -
\eg{scientist} and then generalize to \eg{Mendel} -
\eg{scientist} or \eg{Einstein} - \eg{physicist}, but to learn
from certain categories and entities and then generalize to completely
unseen categories and entities.

It is therefore not surpising that 
without the filtering the model achieves substantially higher scores
(between 2 and 18\% increase), which are arguably the result of learning
which concepts make good categories in general or the specific
relationship between certain categories and the entities that
instantiate them.

\section{Experiment 2: Entity-Based vs.\ Concept-Based
  Instantiation Detection}
\label{sec:exp2-centroid}

Experiment 2 is motivated by the results of the analysis in
Section~\ref{sec:analysis}, namely that the entity embeddings of a
category tend to form a cloud in space, but that the corresponding
category embedding, when computed as a \textit{concept embedding} from
occurrences of the noun denoting the category, is often located
further away. Experiment~1 has demonstrated that it is possible to
learn instantiation to some extent based on the entity
and category embeddings. At the same time, the fact that the best
performance came from a network with two hidden layers and especially the
difficulty of the model dealing with \notinst confounders shows that
it is a rather challenging task. The purpose of
Experiment~2 is to test whether \textit{centroid-based embeddings}, as
defined in Section~\ref{sec:conceptandcategoryreps}, make learning easier.

\subsection{Experimental Setup} \label{sec:method-exp2}

For Experiment~2 we cannot use exactly the same dataset as in Experiment~1,
since some categories have a very small number of entities. We filter
out all categories with less than 5 entities. This reduces the
original set of 577 categories to 159 categories.\footnote{This dataset is also available at \url{http://www.ims.uni-stuttgart.de/data/Instantiation.html}.}

We train and evaluate the same models as in Experiment~1 on the
reduced dataset (using the same anti-memorization methodology and
hyperparameters), with either concept-based or centroid-based
embeddings for categories. The concept-based embeddings are the same
as in Experiment~2.  As for centroid embeddings, they are constructed
as the average of all entity embeddings for this category \eg{in the
  training set}.  Recall that the construction of our datasets
(Section~\ref{sec:dataset} ensures that train, development, and test
sets are disjoint. Thus, for each entity in the test set, it is
guaranteed that its embedding was not used in the construction of any
centroid embedding.  Also, the centroid-based embedding approach,
exactly like the concept-based embedding approach, uses embedding as
the representation for the category that does not change across folds.
Thus, we believe that this setup is as robust to memorization issues
as is the concept-based setup.

\subsection{Results}
\label{sec:results-exp2}

\begin{table}[t]
  \centering
  \caption{  \label{tab:exp2-results}  F$_1$ scores for
    instantiation detection, concept-based vs.\ centroid-based
    category representation (concatenated input). These results are
    computed on the 159 categories with at least 5 entities.}
  \begin{tabular}{l p{1.5cm}  p{1.5cm} p{1.5cm} p{1.5cm} p{1.5cm}}
    \toprule
          & & \multicolumn{2}{c}{Concept-based} &  \multicolumn{2}{c}{Centroid-based}  \\
          \cmidrule(lr){3-4} \cmidrule(lr){5-6}
        Dataset & BL$_{\text{Pos}}$  & \nnone & \nntwo & \nnone & \nntwo \\ \midrule 
	Pos + \inv & 0.67  & \textbf{0.98} & 0.97 & \textbf{0.98} & 0.94	\\
	Pos + \itoi & 0.67 & 0.90 & \textbf{0.91} & \textbf{0.91} & 0.86	\\ 
        Pos + \nir & 0.67  & 0.85 & 0.84 & \textbf{0.90} & 0.89	\\
	Pos + \niid & 0.67 & 0.71 & 0.67 & \textbf{0.79} & 0.75	\\ 
          \midrule
	Pos + \unr & 0.40  & 0.73 & 0.76 & \textbf{0.84} & 0.76	\\ 
	Pos + \unid & 0.40 & 0.51 & 0.65 & \textbf{0.76} & 0.67	\\ 
	\end{tabular}
\end{table}

Table~\ref{tab:exp2-results} shows the results. We focus on the two neural network models, which 
performed well in Experiment~1. The positive class
baseline performs as before, since the class distributions do not
change. The results using a concept-based embedding to represent 
categories (middle columns) can be compared with the corresponding numbers 
in Table~\ref{table:main_results}. The results are rather similar, 
around 1--2\% higher than before. This indicates that the reduced
dataset is comparable in difficulty to the original dataset from
Experiment~1.

Comparing to the centroid-based approach (right-hand columns), we see
that the centroid-based approach does equally well or outperforms the
concept-based one on all variants.  A notable difference from
Experiment~1 is that this time the neural network with a single hidden
layer performs best (see column \nnone), consistently across all
variants. We take this as evidence that the centroid-based
representation requires less transformation of the input, compared to
the concept-based representation.  This is consistent with the
analysis in Section~\ref{sec:analysis}, which showed that entities
are closer in space to their category centroid than to category-denoting nouns.

As for the different variants, Table~\ref{tab:exp2-results} shows that the centroid-based
representation confers gains only for the hardest variants.
On \inv and \itoi, we obtain the same performance,
which makes sense because these variants require the model to
distinguish between entities and categories and deal with similarity
as a confounder.
The centroid-based representation does not provide further help in
this case, presumably for the same reason as above (categories are
more different than entities in the concept-based representation, so
it's easier to distinguish them from entities in the latter case). It
is however noteworthy that it doesn't harm results either.

In contrast, representing categories as centroids of entities yields
big gains for the \notinst and \union cases.
\nir and \unr reach 0.90 and 0.84 performance, respectively, making
them almost comparable to the easier settings (\inv and \itoi have
0.98 and 0.91 F-score).
The hardest setups, \niid and \unid, improve by 8 and 11 points to
0.79 and 0.76 F-score, respectively.

We conclude that, in our setup, recovering the instantiation relation
is much easier from centroid-based representations than from
concept-based representations.

\section{Related Work}
\label{sec:relwork}

\paragraph{Distributional Semantics, Lexical Relations, Entities}
Distributional representations of word meaning have been used widely
to address various linguistic phenomena in computational
linguistics \cite{lenci2008distributional,Turney2010,baroni2011we}.
These include paradigmatic relations like synonym
\cite{landauer1997solution} as well as syntagmatic relations like
predicate-argument relations \cite{erk2010flexible,Baroni2010},
or relations within noun compounds \cite{moldovan2004models}

Among individual lexical relations, hypernymy plays a prominent
role. It was among the first relations to be acquired computationally
by \citet{hearst1992automatic}, although with a pattern-based approach
(see \citet{roller-erk:2016:EMNLP2016} for a recent study combining
both approaches). There is a comprehensive body of work on
distributional modeling of hypernymy
\cite{baroni2011we,lenci2012identifying,roller-erk-boleda:2014:Coling,santus-EtAl:2014:EACL2014-SP,shwartz2016improving}. Hypernymy
is arguably popular because it plays a critical role in many NLP tasks
such as lexical entailment
\cite{Geffet:Dagan:2005,vulic2017hyperlex}, taxonomy detection
\cite{shwartz2017hypernyms}, question-answering
\cite{huang2008question} and cross-lingual inference detection
\cite{upadhyay2018robust}. However, almost all computational
linguistic work on semantic relations focuses on ``classical'' lexical
relations that hold between categories. In contrast, there is hardly
any work specifically on the distributional representation of
entities. These which the exception of \citet{herbelot2015mr} which
analyses the properties specifically of distributional representations
of person names extracted from novels, as well as our own previous
work on extracting entity attributes \cite{gupta2015distributional,gupta2017distributed} and on
instantiation~\cite{boleda2017instances}.



\paragraph{Knowledge Base Population and Completion} A largely
separate line of research, pursued in the semantic web and machine
learning communities, has evolved around the use of structured
knowledge from knowledge bases. NLP applications such as Question
Answering have been attracted by the potential of such knowledge bases
to provide factual and world knowledge about relations among
categories and entities that is arguably necessary for many NLP tasks
such as question answering or semantic inference. This line of
research has seen a huge boost in the last ten years with the
availability of large structured knowledge sources like Wikipedia,
DBPedia \cite{auer2007dbpedia}, Freebase \cite{bollacker2008freebase}
or Yago \cite{suchanek2007yago}. However, even very large knowledge
sources are known to suffer from incomplete coverage
\cite{min2013distant}.  Therefore, the automatic extension and
updating of such resources on the basis of textual information (known
as knowledge base population and completion) has formed its own
research area within which distributional representation play a
prominent role.

\citet{freitas2012querying} discuss the applicability of using 
distributional representations for question-answering systems on 
large-scale commonsense knowledge bases. \citet{socher2013reasoning} 
subsequently show that performance of relational querying models 
improves when entities are represented as an average of their 
constituting word representations built from unsupervised large-scale 
corpora. \citet{bordes2013translating} use such distributional 
representations to indentify the connectivity patterns between entities 
in a multi-relational learning setup on knowledge bases to link entities 
and predict new relations, while \citet{freitas2014distributional} 
introduce a complementary distributional semantic layer to cope with 
semantic approximation and incompleteness of common-sense information. 
\citet{basile2016populating} populate object-location relations in a 
knowledge-base by computing the prototypicality between objects and 
locations extracted from text.

In recent times, the focus has 
shifted towards constructing optimized embeddings for task specific 
application. Link prediction -- connecting two entities through a binary 
relation, is a popular evaluation setup for such embeddings. 
\citet{toutanova2015representing} jointly learn continuous representations 
for entities and textual relations, \citet{nickel2016holographic} use 
distributional representations to generate optimized Holographic embeddings 
to generate compostional embeddings of binary relational data and, 
\citet{trouillon2017knowledge} demonstrate that complex embeddings are more 
optimized as compared to real valued embeddings for link prediction. The 
underlying objective for all of the above is to find expressive yet generalized 
representations which can cope with the ever increasing scalability challenges 
within the scope of this research area. 
The gap that we see in this line of research is that it typically
  induces task-specific embeddings that optimize the performance on
  the task at hand, while we consider vanilla embeddings as they fall
  out of the corpus in the spirit of a generic 'distributional memory'
  \citep{Baroni2010}.  In addition, it is typically evaluated in terms
  of total performance across all relations and rarely provides
  in-depth analyses of individual relations like instantiation.  In
  contrast, we follow a recently emerged direction of research that
  specifically aims at a better understanding of the nature and
  structure of information that is encoded within such representations \cite{lin2015learning,herbelot2015building,gupta2015distributional,xie2016representation,beltagy2016representing}.

\paragraph{Named Entity Recognition and Classification} Named Entity
Recognizion (NER) deals with the identification of named entities in
running and is typically combined with a classification (NEC) of these
entities into categories. Not surprisingly, distributional
representations have found a steady place in both NER and NEC
\cite{xiao2014distributed,gouws2015simple,moreno2017named}.

Traditionally, the recognized categories are rather coarse-grained
(e.g., location, person, organization, other); however, there is a
tendency towards more fine-grained named entity
classification. \citet{Ling:2012:FER:2900728.2900742} were among the
first studies in this direction, performing NEC on a set of 112
classes. \citet{abhishek-anand-awekar:2017:EACLlong} and
\citet{shimaoka-EtAl:2017:EACLlong} present neural architectures for
fine-grained NEC applied to between 47 and 128 classes. 

Our setup is somewhat similar to fine-grained NEC in that we also
  experiment with Named Entities within the bounds of a classification
  setup. However, our set of 577 classes is considerably larger than
  those normally considered in fine-grained NEC.  We also ask our
  models not to disambiguate corpus occurrences of named entities in context,
  but to assess pairs of entities and categories without context, just
  on the basis of embeddings for the entities. We believe that this
  tests the models' ability to acquire the range of possible
  categories for the entities.

\section{Discussion and Conclusion} \label{sec:conclusions}

In this work, we have addressed the \emph{instantiation} relation
between entities and categories, which has received relatively little
attention in linguistics, computational linguistics, and
cognitive science. We started from the observation that the connection
between the theoretical properties of instantiation (e.g., as set
membership in formal semantics) and its practical
operationalization in terms of data-driven meaning representations
remains under-explored.

Building on a large body of previous work that has shown that
distributional semantic representations are a reasonable proxy for
conceptual aspects of
meaning~\cite[among many others]{landauer1997solution,Baroni2010,mikolov2013distributed}, and
on much less work on distributed representations of entities that has
also given encouraging results~\cite{mikolov-yih-zweig:2013:NAACL-HLT,herbelot2015mr,gupta2015distributional}, we have provided a
systematic investigation of the potential of current distributed
representations to capture instantiation.  Our main contributions are
the release of a comprehensive dataset for instantiation, which
enables future work in this area, and the insights obtained through a
set of computational analyses and experiments.

Our findings suggest that the general distinction between categories
and entities, as represented in our data, is easy to recover from
distributional representations, as a simple linear classifier performs
very well in a variety of settings.  In contrast, the instantiation
relationship proper is much more difficult, particularly when using
common nouns as the representation of categories and when the
confounders come from a similar domain: Even with a supervised
approach using non-linear combinations of features (a two-layer neural
network classifier), when controling for memorization it is difficult
to distinguish instantiation from mere semantic relatedness.  The
interference of semantic relatedness is common to lexical semantic
phenomena such as synonymy, hypernymy, or meronymy, as identified in
previous work~\cite{baroni2011we,santus-EtAl:2014:EACL2014-SP,levy-EtAl:2015:NAACL-HLT}.  An additional difficulty in our
case is posed by referential aspects of meaning, such as the
referential ambiguity of names and changes in the world over time.

Our experiments also contribute towards a better understanding of
entities and categories as represented in 
distributional space. A result we did not anticipate is that our entity
embeddings are much more compatible with centroid-based categories, a representation that bases the definition of the category on its
instances, than with concept-based categories, or representations
based on common nouns denoting categories. A possible confounder here is that our entity embeddings are obtained
from names only (\eg{George Washington, Oaxaca, Nile}); by design, other
referential expressions are excluded, and it could be that our results are
due to this design choice. In this case, the observed difference would
be due to the difference in use between proper and common nouns, rather
than between entities and categories represented as common
nouns. Future work should examine this possibility further.

However, our results mesh well with previous work showing that the
so-called Distributional Inclusion Hypothesis for
hypernymy~\citep{Geffet:Dagan:2005} is not borne out for generic
vectors~\citep{roller-erk-boleda:2014:Coling} -- but see
\citet{chang-EtAl:2018:N18-1} for a positive result on a task-optimized
embedding space.
In 
linguistics, hypernymy is traditionally represented in
terms of set inclusion (all dogs, cats, spiders, etc. are animals, so
\eg{animal} is taken to denotes the superset of \eg{dog, cat, spider},
etc.).  Distributional representations are extracted from contexts of
use;
\citet{Geffet:Dagan:2005} translated the set inclusion treatment to distributional
semantics, hypothesizing that the \emph{contexts of use} of, e.g.,
\eg{animal}, would be a superset of the context of use of its
hyponyms, which would be mirrored in an inclusion relation in the
dimensions of the respective distributed representations.  However,
this hypothesis did not hold up empirically, possibly because the word
\eg{animal} is actually used in different contexts than the union of
its hypernyms. This result, as well as our own, suggests that the
identification of common nouns and other content word with ontological
categories, as has been explicitly or implicitly done in much work in
e.g.\ formal semantics and Knowledge Representation, may not be
warranted.  Therefore, our results potentially speak to the broader
issue of the relationship between nouns and other content words, on
the one hand, and ontological categories, on the other. Elucidating
this issue further seems like a fruitful avenue for future work.


Future work could also test whether the centroid-based representation
is also useful for hypernymy detection: Based on our results and the
considerations above, we would expect the average of the hyponyms to
be a more useful representation of the category than the hypernym
itself. Similarly, it should explore the practical implications of our
results (if any) for Natural Language Processing tasks
related to grounding language in the external world, such as Knowledge
Base Population, various forms of Entity Linking, co-reference
resolution, and Referential Expression Generation.

As for instantiation and entity representations, we need to move
beyond the representation of entities via their names, and also
towards ``private'' entities such as my neighbor or the bird I saw
this morning (most work on entities in distributional semantics is on
``public'' entities such as George Washington or Nile; \citet{boleda2017living} is an exception). The problem here is
twofold: On the one hand, distributional methods are extremely
data-hungry, such that it is hard to build meaningful representations
from small samples (this is an area of active research, see
e.g.~\citet{herbelot-baroni:2017:EMNLP2017}). On the other, current
co-reference annotation tools are not reliable enough to build
representations based on all types of referential expressions, as
opposed to relying on names, which can be detected via surface
cues. These are challenges that need addressing if we are to use
distributional methods to better understand the relationship between
entities and categories as expressed in language, and between language
and cognition more generally.

\subsubsection*{Acknowledgments}
AG and SP received funding from Deutsche Forschungsgemeinschaft (DFG)
through Sonderforschungsbereich 732, project D10. GB received funding
from the European Research Council (ERC) under the European Union's
Horizon 2020 research and innovation programme (grant agreement No
715154), the Ramón y Cajal programme (grant RYC-2015-18907), and the
Catalan government (SGR 2017 1575). We gratefully acknowledge the
support of NVIDIA Corporation with the donation of GPUs used for this
research. This paper reflects the authors' view only, and the EU is
not responsible for any use that may be made of the information it
contains.
\begin{flushright}
\includegraphics[width=0.8cm]{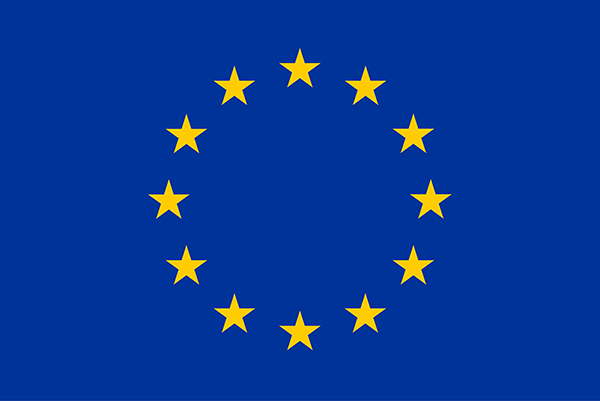}  
\includegraphics[width=0.8cm]{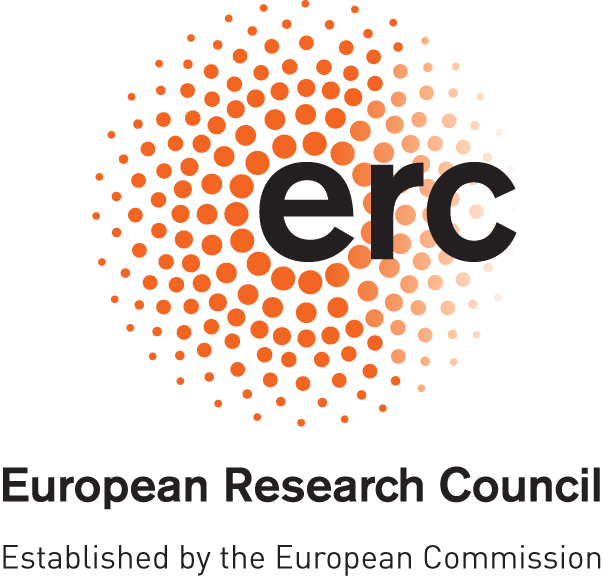} 
\end{flushright}

\starttwocolumn
\bibliographystyle{compling}
\bibliography{instances}

\end{document}